\documentclass[conference]{IEEEtran}
\IEEEoverridecommandlockouts
\usepackage{cite}
\usepackage{amsmath,amssymb,amsfonts}
\usepackage{algorithmic}
\usepackage{algorithm2e}
\usepackage{graphicx}
\usepackage{algorithmic}
\usepackage{graphicx}
\usepackage[utf8]{inputenc}
\usepackage{tikz}
\usepackage{float}
\usepackage[bookmarks=false]{hyperref}
\usepackage{adjustbox}
\usepackage{rotating}

\usepackage{tikz}

\usetikzlibrary{matrix,shapes,arrows,positioning,chains}

\usepackage{setspace}
\begin{document}

\tikzstyle{decision} = [diamond, draw, fill=blue!20, 
    text width=4.5em, text badly centered, node distance=3cm, inner sep=0pt]
\tikzstyle{block} = [rectangle, draw, fill=blue!20, 
    text width=5em, text centered, rounded corners, minimum height=4em]
\tikzstyle{line} = [draw, -latex']
\tikzstyle{cloud} = [draw, ellipse,fill=red!20, node distance=3cm,
    minimum height=2em]

\vspace{-15mm}

\title{GP-RVM: Genetic Programing-based Symbolic Regression Using Relevance Vector Machine
}
\vspace{-15mm}
\author{\IEEEauthorblockN{
${\text{Hossein Izadi Rad}}^{\dagger}$ \thanks{$^{\dagger}$Affiliation: Information and Communication Engineering Department, Graduate School of Information Science and Technology, The University of Tokyo, Tokyo, JAPAN}
}
\IEEEauthorblockA{
izadi@iba.t.u-tokyo.ac.jp}
\and
\IEEEauthorblockN{
${\text{Ji Feng}}^{\dagger}$
}
\IEEEauthorblockA{
feng@iba.t.u-tokyo.ac.jp
}
\and
\IEEEauthorblockN{
${\text{Hitoshi Iba}}^{\dagger}$
}
\IEEEauthorblockA{
iba@iba.t.u-tokyo.ac.jp}
\vspace{-15mm}
}
\vspace{-15mm}

\maketitle
\vspace{-15mm}
\begin{abstract}
This paper proposes a hybrid basis function construction method (GP-RVM) for Symbolic Regression problem, which combines an extended version of Genetic Programming called Kaizen Programming and Relevance Vector Machine to evolve an optimal set of basis functions. Different from traditional evolutionary algorithms where a single individual is a complete solution, our method proposes a solution based on linear combination of basis functions built from individuals during the evolving process. RVM which is a sparse Bayesian kernel method selects suitable functions to constitute the basis. RVM determines the posterior weight of a function by evaluating its quality and sparsity. The solution produced by GP-RVM is a sparse Bayesian linear model of the coefficients of many non-linear functions. Our hybrid approach is focused on nonlinear white-box models selecting the right combination of functions to build robust predictions without prior knowledge about data. Experimental results show that GP-RVM outperforms conventional methods, which suggest that it is an efficient and accurate technique for solving SR. The computational complexity of GP-RVM scales in $O( M^{3})$, where $M$ is the number of functions in the basis set and is typically much smaller than the number $N$ of training patterns. 
\end{abstract}
\vspace{1mm}

\vspace{-2mm}
\section{Introduction}
\vspace{-0.5mm}
In this paper, we address a learning task called symbolic regression (SR) [1-3]. Suppose that there is an unknown real-valued function $f$ whose input can be viewed as a $D$-dimensional vector:
$y = f(x)$
where $x=(x_1,   x_2, \dots,   x_D )^T$ and we are given a set of $N$ observed input-output pairs:
$\{(x_n,y_n )\}_{(n=1)}^N=\{{(x_1,y_1 ),   (x_2,y_2 ), \dots ,   (x_N,y_N)}\}$
where $y_n=f(x_n)$. 
Regression is defined as a set of statistical processes for approximating the relationships between input-output pairs. Symbolic Regression is performed by methods which minimize various error metrics while searching the space of mathematical expressions to estimate the accurate and simple model that best fits the observed dataset [12]. Conventional learning methods such as Artificial Neural Networks (ANNs) [6, 7] and Support Vector Machines (SVMs) [5] has been widely used for solving regression problems. These learning approaches are known as black-boxes, which means the mechanisms behind are difficult to understand and analyze. SR, on the contrary, tries to result in white box models that are clear to interpret. The goal of SR exercise is to determine which of the inputs are the most effective in predicting outputs and identify the input-output relationship [21]. 
One of the significant differences of SR compared to other regression problems is the motivation for finding the exact correct solution. When solving SR, not only the unknown coefficients are optimized, but also the exact formulae that explain the data are determined [8-10]. Linear and nonlinear regression methods try to fit parameters to a given form of an equation. However, SR methods search both the parameters space and the form of equations at the same time, aiming at solving SR form mathematical equations in an easy-to-interpret format. The main reason for choosing SR over other linear and nonlinear regression methods is its interpretability. Such an interpretable formula may not be discovered by other machine learning methods since their primary target is coefficient optimization.

SR is one of the real-world applications of Genetic Programming (GP) [12, 13, 21, 22]. Inspired by Darwin's theory of evolution, GP tries to solve optimization problem by simulating the evolution procedure to evolve computer programs that perform well on a given task. GP is an evolutionary algorithm where many programs are simulated as a cluster of genes that are then evolved based on an evolutionary mechanism. It starts by randomly initializing a population of programs. The population is then repeatedly updated under fitness-based selection, by applying genetic operators until the desired solution is found. GP has been utilized to tackle various problems in different disciplines such as decision making [16-18], pattern recognition [23-25], robotic networks [16-18], bioinformatics [25], data mining, finance [16-18], etc. Despite its successful application in several domains, there are some known issues with GP and related algorithms. For instance, in practice, conventional GP is suffering severely from problems like bloat [26], empirical hyper-parameter tuning, slow speed (as the large size of population), and low success rate [16-18].
\vspace{-0.5mm}
\begin{itemize}
\item{Why Kaizen Programming?}
\end{itemize}
\vspace{-0.5mm}
A controversial issue of GP is its non-determinacy under random search, where the only guide is the pressure on individuals towards better fitness. The algorithm cannot ensure the next iteration to improve, or even avoid to deteriorate because the solutions are randomly modified. Motivated by developing an SR method which requires a relatively lower number of individuals (to accelerate the speed) while still maintaining the quality and interpretability of the solutions, de Melo proposed the Kaizen Programming (KP) [14] in 2014. KP is a cooperative co-evolutionary method which performs automatic feature engineering by using the statistical approach to build models from features generated by GP. KP makes linear models over GP individuals to obtain deterministic results. By keeping on updating a set of best regressors, KP ensures the improvement of solution quality. To improve the solution quality, KP employs feature selection via the null hypothesis significance testing (NHST) based on p-value. KP has an advantage over other similar methods as it relies not only on random evolution but also on a more deterministic and efficient approach to build the model. Rather than focusing on finding a single complete solution, KP concentrates more on how important is each part of the solution and guides the search by dividing the entire task into small ones to achieve high efficiency. The first significant limitation of KP is the introduction of NHST, which raises several controversial issues. For example, the proper selection of a threshold on the significance level, eventually demands prior knowledge. The employment of the linear regression model raises the singularity problem. The singularity happens when duplicated features are used to build the model, and cause the data variance matrix non-invertible.

Motivated by the analytical results, in this paper we extended the feature selection module of KP based on RVM. The proposed method works without performing a hypothesis test, thus requiring no prior knowledge to set the threshold. It also deals with singularity automatically. We implemented the sequential sparse Bayesian learning algorithm to accelerate the training speed of RVM. 
The proposed method is compared to KP and GP on several benchmark functions. Results show that GP-RVM outperforms the other methods, providing more accurate models with fewer evaluations of functions.
The remainder of the paper is organized as follows. In Part 2, we present an overview of the related recent work on solving SR tasks the basic Kaizen Programming workflow. Part 3 describes the method proposed in this work. Experimental results and related topics are discussed in Part 4. Finally, Part 5 summarizes this paper.
 For detailed information about the application of Relevance Vector Machines (RVM) in our work, see the \cite{b31} Chapter 7.

\vspace{-1.5mm}
\section{Previous Works}
\vspace{-0.5mm}
\subsection{Kaizen Programming}
Kaizen Programming is a hybrid method for solving SR based on the Kaizen [14] event with the Plan-Do-Check-Act (PDCA) methodology. In the real world, a Kaizen event is an event where experts propose their ideas and test them to tackle a business issue. Many ideas are then combined to form a complete solution to the issue, which is known as the standard. During the Kaizen process, the PDCA methodology is employed to improve the quality of the current standard, where adjustments are planned, executed, checked and acted. The cycle of PDCA is iterated until the business issue is solved. Since the contribution of each action at each period on resolving the issue is analyzed and evaluated based on specific criteria, more knowledge on the topic is acquired. As a result, the experts learn from the improved information and will avoid useless or harmful actions and control the search process, thus resulting in a better solution.
\begin{itemize}
\item Feature generation module, which is to propose ideas in the shape of regression features.
\item Feature selection module, which is to evaluate and select contributive features.
\item Model generation module, which is to create a complete solution using generated features.
\end{itemize}
For solving the regression problem, a solution can be a mathematical expression, for instance, $2x^3+sin(x)$, which is equivalent to a feature used in a linear regression model. 
In the feature selection module, new ideas are evaluated to generate an actual partial solution, which can be treated as an N-dimensional vector, where N is the number of training data. These partial solutions are combined with old ones existing in the current standard, resulting in a final complete solution composed of partial solutions. Each partial solution is then evaluated to indicate its contribution to solving the problem. Due to the relationship among those partial solutions a measurement considers not only the independent quality of each partial solution but also their dependency is used in this step. The significant difference of conventional evolutionary algorithms with KP is that they work with individuals where each one is a complete solution. Therefore, in KP, individuals collaborate with each other, so the size of the population is smaller compared to conventional evolutionary algorithms, where individuals must compete with each other to survive.

In the model generation module, the complete solution is generated based on the results of the selection module. For example, after several iterations most significant partial solutions are selected from the structure, the coefficients are fitted to the data by performing a Maximum Likelihood Estimation (MLE), which is equivalent to minimize the Root Mean Square Error (RMSE). The evaluation of the quality of the complete solution is also needed. One of the reasons is to estimate if the algorithm gets stuck in a local optimum.

\subsubsection{KP: The Advantages and Disadvantages}
The PDCA cycle plays the role of encouraging experts to propose ideas which are meaningful and useful for solving the problem. It is critical because the criteria for evaluating the contribution replaces the place of fitness on the entire solution. Therefore, the employed methods to perform the second and third modules should be logically connected to each other. For instance, applying the same criteria to choose the features to construct the model, more precisely, the method used to solve the problem and the method employed to evaluate the contribution of the partial solutions are equivalent. Provided that, the feature selection module and the model generation model were unrelated, then the ideas meaningful to the procedure employed in the feature selection module would not be useful to the method that finally solves the problem.


\vspace{-1.5mm}
\subsection{Other Genetic Programming-based Methods for Symbolic Regression}
\subsubsection{Multiple regression genetic programming}
Arnaldo et al. [28] proposed a method which overcomes the limitations of conventional GP by eliminating direct comparison between the output of the complete individual with the target value of the training data. Motivated by the fact that in GP the fitness function is not directly applied to the genotype of individuals, but is instead to the complete phenotype; and thus, resulting in the missing of clear emphasis on the advancing evolution of primitive structures. 
In their experiment, MRGP consistently generated solutions fitter than the result of GP and linear regression.
\vspace{0.5mm}
\subsubsection{Geometric Semantic Genetic Programming (GSGP)}
GSGP [31] concentrates on the introduction of semantic-awareness on genetic operators, which means the awareness of the actual outcome after the operators are applied to individuals. To do so, they formally defined the semantic space as an N-dimensional Euclidean space, where N is the number of training I-O pairs. Each is encoded in the multidimensional metric space as a single point. They proposed several semantic geometric operators which directly search the semantic space, producing offspring that are guaranteed to be at least as fit as the parents. GSGP employs operators which produce geometric offspring to explore the semantic space. Geometric semantic crossover perfectly mixes two parents in the population by constructing an offspring expressed as the (weighted) average of the parents, which is guaranteed to be at least as fit as the less fit of the parents. 
Geometric semantic mutation produces an offspring which is guaranteed to lie in an N-dimensional ball of pre-specified radius centered in the parent. Therefore, it avoids causing colossal migration in the semantic space. 
The primary advantage of GSGP is its relatively higher speed to converge compared to conventional GP. However, the trade-off is its exponentially increasing complexity of the individual, which results in solutions that are theoretically difficult to analyze (black-box) and computationally expensive to evaluate.

%
\vspace{-1mm}
\subsection{Relevance Vector Machine for Symbolic Regression}
Support Vector Machine (SVM) method has been widely used in classification and regression applications. However, machine learning techniques based on support vector machines have some limitations, several of which are explained in \cite{b31} Chapter 7. The output of SVM shows decisions rather than posterior probabilities. These predictive results, are extracted as linear combinations of kernel functions which are focused on training data.
The Relevance Vector Machine \cite{b14} is a Bayesian sparse kernel which has applications in classification and regression. RVM has many qualities similar to SVM. RVM-based solutions avoid fundamental limitations of SVM while resulting in much sparser models. Consequently, RVM corresponds in higher performance on test data. Using RVM for regression we shall find a linear model which results in sparse solutions \cite{b31} (Chapter 3).
Given a train set of input-target pairs ${{\{x_n, t_n\}}_{n=1}^N}$, considering scalar-valued target functions only, it is followed by the standard probabilistic formulation and assume that the targets are samples from the model with additive noise:
\vspace{-3mm}
\begin{equation}\label{eq:2}
\vspace{-1mm}
\hspace{7mm} t_n = \sum_{i=1}^{M}(w_i\phi_i(x_n) + \epsilon_n ) = w^T\phi_i(x_n) + \epsilon_n
\vspace{-1mm}
\end{equation}

where $\epsilon_n$ are independent samples from some noise process which is assumed to be a zero-mean Gaussian with variance $\sigma^2 = \beta^{-1}$ \cite{b31} (Chapter 7.2).
The $\phi_i(x)$ functions are some fixed nonlinear basis functions.
Thus the model defines a conditional distribution for a real-valued target variable $t$, given an input vector $x$, which takes the form:
\vspace{-1mm}
\begin{equation}\label{eq:3}
\vspace{-1mm}
p(t|x, w, \beta) = N(t|w^T\phi(x), \beta^{-1})
\vspace{-0.5mm}
\end{equation}
\vspace{-1mm}
As we assumed the targets $t_n$, the likelihood of the complete dataset can be written as:
\vspace{-3mm}
\begin{equation}\label{eq:4}
\vspace{-1mm}
p(t|x, w, \beta) = \prod_{n=1}^{N} p(t_n|x_n, w, \beta)
\vspace{-1mm}
\end{equation}

Here, RVM adopts a Bayesian perspective and restricts the parameters by introducing a separate hyper-parameter $\alpha_i$ for each of the weight parameters $w_i$ instead of a single shared hyper-parameter\cite{b31} (Chapter 7.2). Thus the weight prior takes the form:
\vspace{-3mm}
\begin{equation}\label{eq:5}
\vspace{-1mm}
p(w|\alpha) = \prod_{n=1}^{M} N(w_i|0, \alpha_i^{-1})
\vspace{-0.3mm}
\end{equation}
\vspace{-0.3mm}
where   $\alpha=(\alpha_1,\dots,\alpha_M)^T$
, and $\alpha_i$ stand for the precision of $w_i$. The introduction of an individual hyper-parameter for every weight is the pivotal feature of the model. The corresponding weight parameters take posterior distributions that are concentrated at 0. Consequently, the basis functions made out of these parameters do not contribute to the model predictions which results in a sparse model.
The posterior over weights is then obtained from the Bayesian rule:
\vspace{-1mm}
\begin{equation}\label{eq:6}
\vspace{-1mm}
p(w| t, x, \alpha, \beta) = N (w| m, \Sigma)
\vspace{-1mm}
\end{equation}
with the posterior mean:
\vspace{-3mm}
\begin{equation}\label{eq:6_1}
\vspace{-1mm}
m = \beta\Sigma\phi^Tt
\vspace{-1mm}
\end{equation} 

 and posterior covariance:
 \vspace{-3mm}
 \begin{equation}\label{eq:6_2}
 \vspace{-1mm}
  \Sigma = {(A+\beta\phi^T\phi)}^{-1}
  \vspace{-1mm}
  \end{equation} 
   where the the $N \times M$ design matrix is defined as $\phi$\cite{b31} (Chapter 7.2, Eq. 7.81). 
The training of relevance vector machine requires to search for posterior hyper-parameter to maximize the marginal likelihood function\cite{b31} (Chapter 7.2, Eq. 7.84):
\vspace{-4mm}
\begin{equation}\label{eq:7}
\vspace{-1.5mm}
p(t| X, \alpha, \beta) = \int p(t | X, w, \beta)p(w|\alpha)dw
\end{equation} 
This integration involves a convolution of two Gaussian functions, it can be easily evaluated by completing the spare to give the log marginal likelihood with:
\vspace{-1mm}
\begin{equation}\label{eq:8}
\vspace{-1mm}
\ln(p(t| X, \alpha, \beta)) = -0.5\{N \ln(2\pi)+ \ln(|C|) + t^TC^{-1}t
\vspace{-1mm}
\end{equation}
\vspace{-1mm}
where
\vspace{-1mm}
\begin{equation}\label{eq:9}
C_{N\times N}=\beta^{-1}I + \phi A^{-1}\phi^T
\vspace{-1mm}
\end{equation}
We try to make the (\ref{eq:8}) as large as possible w.r.t. $\alpha$ and $\beta$ variables. Therefore, we set the derivatives of marginal likelihood to zero to obtain:
\vspace{-1mm}
\begin{equation}\label{eq:10}
\alpha_i^{new} = \cfrac{\gamma_i} {m_i^2} \hspace{10mm}
(\beta^{new})^{-1} = \cfrac{||t-\phi m||^2}{N-\sum _i\gamma _i}
\vspace{-2mm}
\end{equation}

where the $i^{th}$ element of the posterior mean $m$ is $m_i$ defined by (\ref{eq:6_1}). \cite{b31} (Section 3.5.3) defines $\gamma_i$ as a measure on how well the $w_i$ parameter is determined:
\vspace{-1mm}
\begin{equation}\label{eq:12}
\gamma _i = 1 - \alpha _i \Sigma _{ii}
\vspace{-1mm}
\end{equation}
where $\Sigma$ is defined in (\ref{eq:6_2}). 

The learning algorithm starts with initializing $\alpha$ and $\beta$, and then calculating the mean and covariance of the posterior using (\ref{eq:6_1}) and (\ref{eq:6_2}), respectively. The learning algorithm proceeds by repeating the equation of (\ref{eq:10}), accompanied by updating the posterior statistics, until reaching the maximum of repeats or minimizing all of the precision parameters, i.e. $\alpha_i$s, until they get smaller than a lower-bound (\ref{eq:6_1}). [32] (Section 7.2.2).
The optimization found in practice, drives a proportion of the hyper-parameters $\alpha_i$ into large (theoretically infinite) values, as a result, the weight parameters $w_i$ related to hyper-parameters have posterior distributions with mean and variance both zero. Accordingly, those parameters as well as corresponding basis functions $\Phi_i (x)$ should be eliminated from the model and play no role in predicting new inputs [15].
\vspace{-2.5mm}
\section{Proposed Method}
\subsection{An Efficient Hybrid GP-based Approach Using Relevance Vector Machine}
\vspace{-1mm}
This section introduces a new method to SR, based on the Sparse Bayesian kernel method, which integrates a GP-based search of tree structures, and a Bayesian parameter estimation employing automatic relevance determination (ARD). To overcome GP's significant difficulty, we extend the work of KP in [14] by improving the feature selection with ARD, using RVM. Because RVM requires no prior knowledge to set a threshold and deals with singularity automatically, the proposed method also overcomes the limitations of KP. Figure 1 shows the basic flowchart of the proposed method. We explain three key steps which are shown with a star mark $\bigstar$ after the subsequent parts this chapter.
To overcome the limitations of KP, we first analytically introduce regularization terms over the data variance matrix to avoid the singularity. The regularized matrix can be considered as a posterior covariance matrix. It means that the singularity problem can be avoided if we generalize the linear regression model into a Bayesian linear regression model. In that case, there would be distributed precision parameters for each linear coefficient. In other words, the generalized model is equivalent to the well-known RVM [15]. Also, the generalized model no longer depends on the hypothesis test. Therefore, it does not require any prior knowledge to set a proper threshold for feature selection. 

Considering the analytical results, in this paper we introduce a hybrid method for solving SR problem based on RVM. RVM is widely used as a kernel technique although there is no restriction for applying it to any model expressed as a linear combination of feature functions. In our proposed method, GP individuals are used for generating features. A preprocessing is applied for extracting all subtree functions in each tree-formed GP individual, to ensure the maximum usage of provided resources. These functions are further used to build linear models together with features existing in the current model. Different from the statistical methods used in KP, RVM is used to select proper features. A sequential training algorithm [19] is employed to perform the optimization task of RVM which improves training speed significantly. Optimized results are finally checked under the fitness criterion to ensure the improvement of model quality and avoid being stuck in a local optimum. We use the adjusted R-squared ($Adj.R^2$) [20] rather than RMSE as the fitness criterion for model comparison and for reducing overfitting by controlling the complexity of the selected model.

\begin{figure}
\vspace{-6mm}
	\centering
	\includegraphics[width=3.1in]{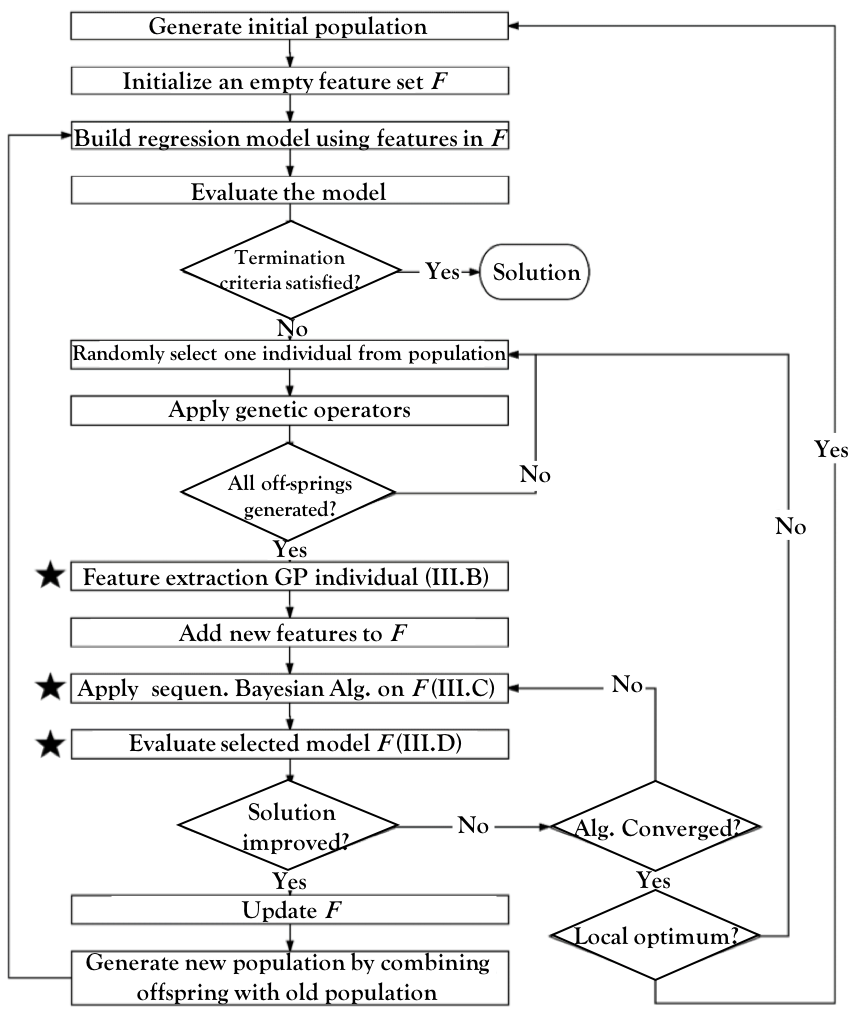}
	\vspace{-4mm}
	\caption
	{Flowchart of the proposed method.}
	\vspace{-6mm}
\end{figure}

\vspace{-2mm}
\subsection{Feature Generation from GP Individuals}
\vspace{-1mm}
A sufficient number of candidate features must be produced in the first place to ensure the quality of the regression model. In the experiment, we found that the variety of features it utilizes strongly influences the performance of the proposed method. On the other hand, the lack of candidate features will decrease the increasing speed of fitness and increase the risk of being stuck in a local optimum. As we no longer consider the individuals as solutions, the primary duty of KP is to generate feature functions. Empirically, we found that a proper feature function usually exists in a subtree of the individual. However, as the mapping from a tree to a function is complicated, the final behavior of the tree can be very different. If we only use the entire tree as one feature function, then all the subtrees are wasted, and it is difficult for them to reappear in the future generation again. One approach is to increase the number of individuals, but it still brings more subtrees to be wasted. So, the proper solution is to do a traversal for every single tree to extract all its subtree functions. The training of RVM will be difficult to converge on a repeated basis or even fail due to the singularity. As a result, in practice, the python computational algebra library (SymPy) is used to transform a primitive tree into a symbolic expression, which will simplify the function (thus reduce the next evaluation complexity) and check the uniqueness of them. As an expression can frequently appear along the process, an LRU (Least Recently Used) cache is implemented to save the evaluation result of an evaluated expression.

In short, it takes two steps for a GP primitive tree to construct candidate functions. First, the entire tree will be traversed to yield all its subtree functions. Second, all the subtrees will be transformed into symbolic expressions and filtered, resulting only different functions.
\vspace{-1mm}
\subsection{Sequential Sparse Bayesian Learning Algorithm}
\vspace{-1mm}
After the construction of candidate functions, the next step is to put them into RVM for training \cite{b31} \cite{b18}. In practice, we use another approach to solving the optimization problem which improves training speed significantly. The problem is to determine the hyper-parameters $\alpha$ and $\beta$:
\vspace{-1mm}
\begin{equation}
\vspace{-1mm}
\alpha^*, \beta^*= argmax_{\alpha, \beta}  (\ln p(t|X, \alpha, \beta))
\vspace{-1mm}
\end{equation}
\vspace{-2mm}

\vspace{-4mm}
\begin{algorithm}
    \underline{Sequential Sparse Bayesian Learning Algorithm}\;
      {
1. Initialize $\beta$, select a candidate function $\Phi_1$, set its $\alpha_1$. Set the other $\alpha_{j\neq1}$ to $\infty$.\\
2. Evaluate the posterior of $w$, along with $q_i$, $s_i$ for all basis functions.\\
3. Select a candidate function $\Phi_i$.\\
4. If $q_i^2\geq s_i$  and $\alpha_i \leq \infty$, update $\alpha_i$.\\
5. If $q_i^2\geq s_i$  and $\alpha_i=\infty$, include $\Phi_i$ to the model, initialize its $\alpha_i$.\\
6. If $q_i^2\leq s_i$, remove $\Phi_i$ from the model, set $\alpha_i=\infty$.\\
7. Update $\beta$. Repeat 2 until convergence.
      }
      \vspace{1mm}
    \caption{Sequential Sparse Bayesian Learning Algorithm}
\end{algorithm}
\vspace{-5mm}
\subsection{Model Selection}
\vspace{-1mm}
The principal disadvantage of RVM is that the training involves optimizing a non-convex function. During the training phase, many distinct models with a different number of active functions will be generated. In general, the model with more active functions behaves better on the training set. However, a simpler one is preferred if the difference of training error between two models is small. The complexity of RVM scales to the number of active bases, in the experiments we found that the speed of the proposed method is significantly influenced by the number of functions kept in the record. As a result, it is essential to compare the models produced in the training phase and select the most straightforward model among those which exceed a measurement of quality.
In [14], the adjusted $R^2$ is used for model comparison, as it considers not only the fitness value but also the number of active functions compared to the amount of training data. The adjusted $R^2$ is calculated using the following formula:
\vspace{-1mm}
\begin{equation}
Adj.R^2=R^2-(1-R^2)p/(N-p-1)
\vspace{-2mm}
\end{equation}
where N is the number of training patterns, p is the number of active functions in a model and $R^2$ is the constant of determination which is given by:
\vspace{-2mm}
\begin{equation}
\vspace{-2mm}
R^2=1-(\sum_{n=1}^{N}((t_n-y_n )^2 )/(\sum_{n=1}^{N}(t_n-\bar{t}))
\vspace{-1mm}
\end{equation}
where $\bar{t}=1/N \sum_{n=1}^{N}t_n$ , and $y_n$ is the $n^th$ entry of the output vector of a model.
In the proposed method, the $Adj. R^2$ is used as the criterion to select the best model from the converged models. Compared to RMSE, the advantage of Adj. $R^2$ is that it is scale-free, w.r.t. the absolute target values of the problem and the result of this statistic is in the range (0, 1) where 1.0 means a perfect fit, $ \ge$ 0.99 means a high-quality model and $ \ge$ 0.95 means a medium-quality model.
\vspace{-0.5mm}
\section{Experimental Results}
This chapter presents a comparison of our proposed method against the conventional GP and the KP shown in the literature [14] to solve symbolic regression benchmark functions [33, 34]. The experiments demonstrate that our method can outperform them, providing high-quality solutions for both training and testing sets.
\vspace{-2mm}
\subsection{Benchmarks}
\vspace{-1mm}
Several $keijzer$ benchmarks in Table I [33] for the first experiment and the $nguyen$ benchmarks in Table II [34] for the second experiments which were chosen to compare our method with the results presented in [14]. Similar to [14], $c$ randomly sampled points from the uniform distribution confined in the range $[a,b]$ are noted as $U[a,b,c]$. Points confined the range $[a,b]$ with successively equal intervals $c$ are noted as $E[a,b,c]$. 
\vspace{-1mm}
\subsection{Configuration}
The Distributed Evolutionary Algorithms in Python 3.6.4 (DEAP) [35] is used to implement the GP-related parts of our method, which are the population generation based on genetic operators. The Python library for symbolic mathematics (SymPy) is used as the computational algebra library. The Python library for numeric computations (NumPy) is used for the numerical calculation in RVM and evaluation of functions in the method. Table III. shows the configuration and parameters setting for the $keijzer$ experiment. Table IV. is the configuration and setting for the $nguyen$ benchmarks. Missing terms of Table IV could be found in Table III.

\begin{table}[]
\centering
\vspace{-9mm}
\caption{$keijzer$ Benchmark Functions}
\label{Table1}
\vspace{-3mm}
\begin{tabular}{llll}

\hline

Function                                 & Training data   & Testing data      &  \\
\hline
$keizer1=0.3xsin(2\Pi x)$                     & E{[}-1,1,0.1{]} & E{[}-1,1,0.001{]} &  \\ \hline
$keizer2=0.3xsin(2\Pi x)$                     & E{[}-2,2,0.1{]} & E{[}-2,2,0.001{]} &  \\ \hline
$keizer3=0.3xsin(2\Pi x)$                     & E{[}-3,3,0.1{]} & E{[}-3,3,0.001{]} &  \\ \hline
$keizer6=\sum_{i=1}^{x} 1/i$ 	& E{[}1,50,1{]}   & E{[}1,120,1{]}    &  \\ \hline
$keizer7=\ln x$                             & E{[}1,100,1{]}  & E{[}1,100,0.1{]}  &  \\ \hline
$keizer8=\sqrt{x}$                              & E{[}1,100,1{]}  & E{[}1,100,0.1{]}  &  \\ \hline
$keizer9=arcsinh(x)$                       & E{[}1,100,1{]}  & E{[}1,100,0.1{]}  &  
\vspace{-7mm}
\end{tabular}
\end{table}

\begin{table}[]
\centering
\vspace{-9mm}
\caption{$nguyen$ Benchmark Functions}
\label{Table2}
\vspace{-3mm}
\begin{tabular}{lll}
\hline
Function                                                                                                      & Train / Test &  \\ \hline
$nguyen1=x^3+x^2+x$                                                             & U{[}-1,1,20{]}          &  \\ \hline
$nguyen2=x^4+x^ 3+x^ 2+x$                                         & U{[}-1,1,20{]}          &  \\ \hline
$nguyen3=x^ 5+x^ 4+x^ 3+x^ 2+x$                     & U{[}-1,1,20{]}          &  \\ \hline
$nguyen4=x^ 6+x^ 5+x^ 4+x^ 3+x^ 2+x$ 		& U{[}-1,1,20{]}          &  \\ \hline
$nguyen5=sin(x^ 2 ) cos(x)-1$                                                                    & U{[}-1,1,20{]}          &  \\ \hline
$nguyen6=sin(x)+sin(x+x^ 2)$                                                                     & U{[}-1,1,20{]}          &  \\ \hline
$nguyen7=\log(x+1)+\log(x^ 2+1)$                                                                   & U{[}0,2,20{]}           &  \\ \hline
$nguyen8=\sqrt{x}$                                                                                                   & U{[}0,4,40{]}           &  \\ \hline
$nguyen9=sin(x)+sin(y^ 2)$                                                                       & U{[}-1,1,100{]}         &  \\ \hline
$nguyen10=2sin(x)cos(y)$                                                                                        & U{[}-1,1,100{]}         & 
\vspace{-3mm}
\end{tabular}
\end{table}

For comparison, the proposed method and the other methods in the KP [14] were configured as shown in Table II. Configurations regarding KP, GP50, and GP500 are referenced from [14]. The maximum generations are set to result in a balance of the total number of individuals used among all methods. In the first experiment, the execution of each process will be terminated if the maximum generation is reached or the current global solution has higher fitness than 0.99999. In the second experiment, the termination condition is set to the maximum evaluation of nodes achieved. The termination will be considered as a failure if, in any one of the training I-O case, the absolute error is more significant than 0.01. Our method is executed 50 times in the first experiment, and 100 times in the second experiment on each benchmark.

\begin{table}[]
\centering
\caption{Configuration and Parameter Setting for $keijzer$ Benchmark}
\label{Table3}
\vspace{-2mm}
\begin{tabular}{ll}
\hline
Parameter                  & Value                                                                                   \\ \hline
 & GP-RVM / KP/ GP50/ GP500                                                                                             \\ \hline
 \hline
Population size            & 8/ 8/ 50/ 500                                                                           \\ \hline
Max. generations           & 2000/ 2000/ 500/ 50                                                                     \\ \hline
Crossover probability      & 1.0/ 1.0/ 0.9/ 0.9                                                                     \\ \hline
Crossover operator         & One-point                                                                               \\ \hline
Mutation probability       & 1.0/ 1.0/ 0.05/ 0.05                                                                    \\ \hline
Mutation operator          & 90\% Uniform \& 10\% ERC/ \\ &Uniform/ Uniform                                                                                               \\ \hline
Max. depth                 & 15/ 2/ 15/ 15                                                                           \\ \hline
Non-terminals              & $+,\times,1/n,-n,\sqrt{|n|}$                                                            \\ \hline
Terminals                  & $x, N(\mu=0,\sigma=5)$                                    \\ \hline
Fitness                    & $Adj.R^ 2/Adj.R^ 2/R^ 2/R^2$ \\ \hline
Stopping criteria          & Max. gen. or fitness\textgreater0.99999 \\ \hline
Runs                       & 50/ 50/ 50/ 50                   
\vspace{-7mm}                                         
\end{tabular}
\end{table}

\begin{table}[]
\vspace{-3mm}
\centering
\vspace{-3mm}
\caption{Configuration and Parameter Setting for $nguyen$ Benchmark}
\label{Table4}
\vspace{-3mm}
\begin{tabular}{ll}
\hline
Parameter                  & Value \\ \hline
& GP-RVM/ KP/ GP50/ GP500 \\ \hline
\hline
Population size            & 4/ 8/ 50/ 500                                               \\ \hline
Max. node eval.   & 2000                                                        \\ \hline
Non-terminals              & $+,-,\times,\div, \sin,\cos, \exp, \log$ \\      \hline                                                     
Terminals                  & $x, Constant 1$ ($ng.1$  to $ng. 8)$,\\ & $y$ $  (ng. 9$  and  $ng. 10)$ \\ \hline
Runs                       & 100/ 100/ 100/ 100   
\vspace{-7mm}                                      
\end{tabular}
\end{table}

\vspace{-1.5mm}
\subsection{Results}
\vspace{-1mm}
The first experiment is on $keijzer$ benchmarks and is for comparing the model quality and computational complexity. The training results are shown in Table V, and the testing results are shown in Table VI. The number of function evaluations (NFEs) is posted for comparison of computational complexity. The statistical results are collected based on 50 times of trails. The second experiment is on $nguyen$ benchmark functions. It uses the same interval for training and testing, but the sets of points are distinct. The objective is to result in the absolute error of each training I-O pair is smaller than 0.01 within a finite number of node evaluations. The result of our method is shown in Table VII, and the comparison with other methods is referenced and shown in Figure 2. The statistical results are collected based on 100 times of trails.

\begin{table*}[]
\vspace{-10mm}
\caption{Training Result on $keijzer$ Benchmark}
\label{Table5}
\vspace{-3mm}
\begin{tabular}{llllllllllllll}
  \cline{3-5} \cline{9-11}
F    & Stat. &            & GP-RVM   &       &            & KP       &       &            & GP500    &       \\ \hline
     &       & $Adj. R^2$ & RMSE     & NFEs  & $Adj. R^2$ & RMSE     & NFEs  & $Adj. R^2$ & RMSE     & NFEs  \\ \cline{3-5} \cline{9-11}
     & Min   & 1.00E+00   & 8.18E-05 & 532   & 1.00E+00   & 1.93E-06 & 328   & 4.12E-01   & 8.00E-02 & 26700 \\ \hline
$k1$ & Med.  & 1.00E+00   & 4.14E-05 & 5311  & 1.00E+00   & 2.21E-04 & 9492  &4.14E-01   & 8.97E-02 & 30450 \\ \hline
     & Max   & 1.00E+00   & 2.06E-05 & 11214 & 1.00E+00   & 2.28E-02 & 32260 & 5.34E-01   & 8.99E-02 & 32170 \\ \hline
     & Min   & 9.95E-01   & 8.49E-04 & 1052  & 1.25E-01   & 5.42E-04 & 2880  & 5.50E-02   & 2.17E-01 & 28460 \\ \hline
$k2$ & Med.  & 9.98E-01   & 9.04E-03 & 6654  & 8.99E-01   & 6.75E-02 & 32190 & 1.39E-01   & 2.23E-01 & 30550 \\ \hline
     & Max   & 1.00E+00   & 1.09E-02 & 16588 & 1.00E+00   & 2.19E-01 & 32380 & 1.84E-01   & 2.34E-01 & 32310 \\ \hline
     & Min   & 9.56E-01   & 3.89E-02 & 11130 & 1.26E-01   & 9.98E-02 & 32100 & 3.26E-02   & 3.35E-01 & 28300 \\ \hline
$k3$ & Med.  & 9.81E-01   & 9.75E-02 & 14811 & 3.83E-01   & 2.65E-01 & 32170 & 7.05E-02   & 3.51E-01 & 30300 \\ \hline
     & Max   & 9.92E-01   & 2.42E-01 & 17870 & 9.13E-01   & 3.36E-01 & 32260 & 1.53E-01   & 3.58E-01 & 31950 \\ \hline
     & Min   & 1.00E+00   & 4.26E-04 & 7     & 1.00E+00   & 6.99E-16 & 56    & 1.00E+00   & 9.90E-01 & 501   \\ \hline
$k6$ & Med.  & 1.00E+00   & 4.26E-04 & 19    & 1.00E+00   & 9.22E-16 & 56    & 1.00E+00   & 9.90E-01 & 501   \\ \hline
     & Max   & 1.00E+00   & 1.27E-04 & 104   & 1.00E+00   & 1.18E-14 & 14510 & 1.00E+00   & 9.90E-01 & 501   \\ \hline
     & Min   & 1.00E+00   & 1.32E-03 & 7     & 1.00E+00   & 2.96E-06 & 56   & 9.96E-01   & 1.05E-01 & 28590 \\ \hline
$k7$ & Med.  & 1.00E+00   & 2.26E-03 & 8     & 1.00E+00   & 4.61E-04 & 80    & 9.98E-01   & 1.81E-01 & 30660 \\ \hline
     & Max   & 1.00E+00   & 6.83E-03 & 43    & 1.00E+00   & 1.78E-02 & 224   & 9.99E-01   & 2.51E-01 & 32480 \\ \hline
     & Min   & 1.00E+00   & 1.03E-06 & 7     & 1.00E+00   & 7.91E-16 & 56    & 1.00E+00   & 0.00E+00 & 501   \\ \hline
$k8$ & Med.  & 1.00E+00   & 5.61E-05 & 7     & 1.00E+00   & 3.50E-15 & 56    & 1.00E+00   & 0.00E+00 & 501   \\ \hline
     & Max   & 1.00E+00   & 6.08E-03 & 38    & 1.00E+00   & 1.34E-02 & 160   & 1.00E+00   & 0.00E+00 & 501   \\ \hline
     & Min   & 1.00E+00   & 8.08E-04 & 7     & 1.00E+00   & 9.77E-06 & 56    & 9.69E-01   & 1.12E-01 & 25500 \\ \hline
$k9$ & Med.  & 1.00E+00   & 5.41E-03 & 7     & 1.00E+00   & 1.59E-03 & 88   &  9.99E-01   & 1.35E-01 & 30300 \\ \hline
     & Max   & 1.00E+00   & 9.19E-03 & 22    & 1.00E+00   & 2.91E-02 & 248   & 9.99E-01   & 7.70E-01 & 32050
\end{tabular}
\vspace{-5mm}
\end{table*}

\begin{table}[]
\centering
\vspace{-2mm}
\caption{RMSE Testing Result on $keijzer$ Benchmark}
\label{Table6}
\vspace{-3mm}
\begin{tabular}{lllll}
\hline
Func.    & Stat.  & GP-RVM   & KP       & GP 500   \\
\hline
& min    & 7.10E-03 & 1.07E-05 & 7.88E-02 \\ \hline
$keijzer1$          & median & 2.56E-02 & 4.36E-04 & 8.20E-02 \\ \hline
         & max    & 1.00E-02 & 5.74E+06 & 1.30E+00 \\ \hline
 & min    & 9.77E-03 & 6.86E-04 & 2.09E-01 \\ \hline
$keijzer2$         & median & 6.76E-02 & 6.85E-02 & 2.22E-01 \\ \hline
         & max    & 1.57E+00 & 8.23E+00 & 1.00E+00 \\ \hline
 & min    & 1.10E-02 & 9.67E-02 & 3.41E-01 \\ \hline
$keijzer3$         & median & 5.02E-01 & 2.07E+00 & 3.53E-01 \\ \hline
         & max    & 1.30E+01 & 9.98E+01 & 2.71E+00 \\ \hline
 & min    & 6.87E-03 & 5.49E-16 & 9.96E-01 \\ \hline
$keijzer6$         & median & 9.71E-03 & 8.81E-16 & 9.96E-01 \\ \hline
         & max    & 2.22E-02 & 8.03E-14 & 9.96E-01 \\ \hline
 & min    & 1.59E-03 & 3.89E-04 & 1.06E-01 \\ \hline
$keijzer7$         & median & 1.83E-03 & 1.27E-02 & 1.71E-01 \\ \hline
         & max    & 4.42E-02 & 4.49E+02 & 5.50E+08 \\ \hline
 & min    & 1.03E-06 & 1.23E-15 & 0        \\ \hline
$keijzer8$         & median & 1.78E-04 & 6.08E-15 & 0        \\ \hline
         & max    & 2.68E-03 & 1.63E+00 & 0        \\ \hline
 & min    & 8.18E-04 & 4.74E-03 & 1.39E-01 \\ \hline
$keijzer9$         & median & 4.55E-03 & 1.03E-01 & 1.51E-01 \\ \hline
         & max    & 9.64E-03 & 3.82E+09 & $\infty$      
\end{tabular}
\vspace{-8mm}
\end{table}

%

\begin{sidewaystable}
\centering
\vspace{-22mm}


\label{Table7}
\caption{Results (average of 100 trials) of $nguyen$ benchmark functions. Raw fitness is sum of absolute error on all fitness cases.}
\begin{tabular}{llllllll}
\hline
Problem  & Max. Error & Raw Fitness & RMSE Training & Func. Eval. & Node Eval. & RMSE Testing & Succ. Runs \\
\hline
$nguyen1$  & 0.00283       & 0.01823     & 0.00112       & 21.14       & 20.63      & 0.05112      & 100        \\ \hline
$nguyen2$ & 0.00479       & 0.02357     & 0.00135       & 32.63       & 27.56      & 0.10625      & 100        \\ \hline
$nguyen3$  & 0.00399       & 0.02711     & 0.00172       & 45.73       & 36.68      & 0.83062      & 100        \\ \hline
$nguyen4$  & 0.00406       & 0.02746     & 0.00174       & 33.41       & 54.32      & 0.49868      & 100        \\ \hline
$nguyen5$  & 0.00057       & 0.00396     & 0.00025       & 14.39       & 26.62      & 0.36413      & 100        \\ \hline
$nguyen6$  & 0.00264       & 0.0188      & 0.00118       & 18.17       & 29.14      & 0.099        & 100        \\ \hline
$nguyen7$  & 0.0022        & 0.01473     & 0.00093       & 8.07        & 14.04      & 0.28685      & 100        \\ \hline
$nguyen8$  & 0.00145       & 0.00948     & 0.0006        & 17.26       & 29.66      & 0.1638       & 100        \\ \hline
$nguyen9$  & 0.00352       & 0.09064     & 0.00114       & 75.29       & 177.51     & 0.70064      & 100        \\ \hline
$nguyen10$ & 0.00442       & 0.10199     & 0.00133       & 129.54      & 311.68     & 0.20869      & 100       
\end{tabular}
\end{sidewaystable}

\vspace{-2mm}
\subsection{Discussion}
\vspace{-1mm}
The first experiment is to verify that if our method could outperform regular GP, and its predecessor, KP. For seven $keijzer$ functions, the training results show that our approach achieved the highest values of fitness for two functions ($keijzer$ 2 and 3), with five ties to KP ($keijzer$ 1, 6, 7, 8, 9) and 2 ties to GP ($keijzer$ 6 and 8). However, the RMSE value of our method is more significant when ties. Our way found high-quality models ($median$ of fitness $\geq$0.99 for six functions, whereas KP found five and GP (50 and 500) found models with such quality for only four functions.
To explain the results, we learn that our method performs as fast as KP and GP for simple problems and it performs better when dealing with complex functions ($keijzer$ 2 and 3), where KP can find models with low or poor quality and GP can only find poor results. For these functions, our method requires a much smaller NFEs ($median$) than the other methods. It should be noted that, when dealing with easy targets ($keijzer$ 6, 7, 8, 9), the models found by our method are sparse and rapidly meet the stopping criteria (fitness$\geq$0.99999), which explains the RMSE value of our method is more significant as the $Adj.R^2$ rewards sparse models.
Concerning the testing results, we see that the min and $median$ of RMSE of our method are somewhat larger than KP ($keijzer$ 1, 6, 7, 8, 9). However, the maximum errors of GP-RVM are far smaller than KP in $keijzer$ 1, 2, 3, 7, 8, 9, which explains that the results of our method are more stable than the others, in other words, more robust on the testing set. In practice, the robustness of the testing set is much more important than a small difference in error.

The second experiment is to test curve-fitting tasks on $nugyen$ benchmark functions where the maximum absolute error of any of the fitness cases should be lower than 0.01. Table VII shows the averaged results of our method. Like KP, GP-RVM succeeds to fit target curves as the RMSE for training and testing remain low values for every test.
Regarding the efficiency, we notice that the number of objective function and node evaluations also remain low values for every test, which means that our method is faster in finding the correct result. Part of the reason is the implementation of cache, but more importantly, is the sparse kernel method employed by our method results in fast convergence. Figure 2. compares the results of our method with KP and other techniques working on the same curve-fitting problem. The performance of our method is as accurate as KP, both yeilding up 100\% successful runs. Both GP-RVM and KP outperform other methods.

\vspace{-0.5mm}
\section{Conclusion}
\vspace{-1.5mm}
\subsection{Summary}
\vspace{-0.5mm}
In this paper we focus on the topic of symbolic regression (SR), using symbols of functions and variables to construct solutions. Expanded search space and the reduced amount of prior knowledge is the main reason for choosing SR instead of other machine learning models. To perform SR, Genetic Programming (GP) is a representative method, which is a population-based evolutionary algorithm. Limitations including a blind search of GP attracted many works to remove them. Kaizen Programming (KP) is a successful one among those works, which combines linear regression modeling and hill-climbing approach to guides the search. Though its success, problems including threshold setting and singularity raise.
We analytically found that the singularity problem can be solved by generalizing the linear model of KP. Moreover, if we take a Bayesian treatment to estimate the generalized model, we can avoid using hypothesis testing to perform feature selection, where the threshold setting problem solved simultaneously. The generalized model, found in literature, is known as the relevance vector machine (RVM).
Motivated by the analytical results, in this work we extended the feature selection module of KP based on RVM. The proposed method works without performing a hypothesis test, thus requiring no prior knowledge to set the threshold. It also deals with singularity automatically. We implemented the sequential sparse Bayesian learning algorithm to accelerate the training speed of RVM. The proposed method is compared to KP and GP on several benchmark functions. Results showed that it outperforms the other methods, providing more accurate models with fewer functions evaluations.

\begin{sidewaysfigure}
\vspace{100mm}
    \begin{tikzpicture}[scale=3.5]
    \includegraphics[width=9.5in]{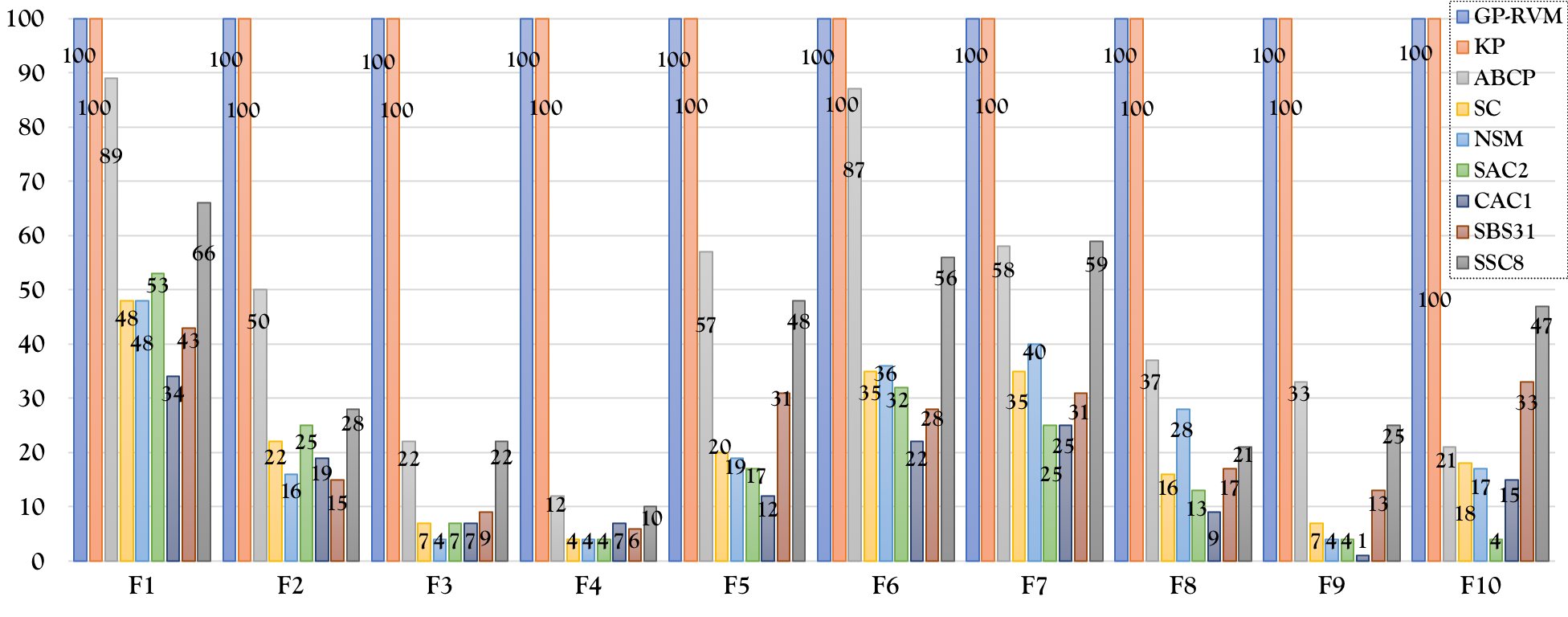}
    \end{tikzpicture}
\caption{Number of successful runs using Nguyen benchmark functions (F1-F10) }
    \label{fig:awesome_image}
\end{sidewaysfigure}


\vspace{-3mm}
\begin{thebibliography}{36}
\vspace{-1mm}
\bibitem{b0}de Melo, V.V. and Banzhaf, W. 2016. Kaizen programming for feature construction for classification. In Genetic Programming Theory and Practice XIII, pp. 39-57, Springer.

\bibitem{b1} Peng, Y., Yuan, C., Qin, X., Huang, J. and Shi, Y. 2014. An improved gene expression programming approach for symbolic regression problems. Neurocomputing, 137, pp.293-301.

\bibitem{b2} Tang, F., et. al. 2015. Discovery scientific laws by hybrid evolutionary model. Neurocomputing, 148, pp.143-149.

\bibitem{b3} Icke, I. and Bongard, J.C. 2013, June. Improving genetic programming based symbolic regression using deterministic machine learning. In Evolutionary Computation IEEE Congress on pp. 1763-1770.

\bibitem{b4}Cristianini, N. and Shawe-Taylor, J., 2000. An introduction to support vector machines and other kernel-based learning methods. Cambridge university press.

\bibitem{b5}Shafaei, M. and Kisi, O.,2017. Predicting river daily flow using wavelet-artificial neural networks based on regression analyses in comparison with artificial neural networks and support vector machine models. Neural Computing and Applications, 28(1), pp.15-28.

\bibitem{b6}Deklel, A.K., Saleh, M.A., Hamdy, A.M. and Saad, E.M. 2017, March. Transfer learning with long term artificial neural network memory (LTANN-MEM) and neural symbolization algorithm (NSA) for solving high dimensional multi-objective symbolic regression problems. In Radio Science Conference, 2017 34th National (pp. 343-352). IEEE.

\bibitem{b7}Austel, V., et. al. 2017. Globally Optimal Symbolic Regression. arXiv preprint arXiv:1710.10720. 

\bibitem{b8}Michael S., and Hod L. 2009. Distilling Free-Form Natural Laws from Experimental Data. Science Col. 324, Issue 5923, pp.81-85.

\bibitem{b9}Michael S., and Hod L. 2013. Eureqa software (version 0.98 beta). Nutonian, Somerville, Mass, USA.

\bibitem{b10}Michael S., and Hod L. 2013. Symbolic Regression of Implicit Equations. Genetic Programming Theory and Practice VII. Genetic and Evolutionary Computation. Springer, Boston, MA.

\bibitem{b11}Koza, J.R., 1994. Genetic programming as a means for programming computers by natural selection. Statistics and computing, 4(2), pp.87-112.

\bibitem{b12}Banzhaf, W., Nordin, P., Keller, R.E. and Francone, F.D. 1998. Genetic Programming: An Introduction: On the Automatic Evolution of Computer Programs and Its Applications. Morgan Kaufmann Publishers. Inc., Heidelberg and San Francisco CA.

\bibitem{b13}de Melo, V.V. 2014, July. Kaizen programming. In Proceedings of the 2014 Annual Conference on Genetic and Evolutionary Computation (pp. 895-902). ACM.

\bibitem{b14}Tipping, M.E. 2001. Sparse Bayesian learning and the relevance vector machine. Journal of machine learning research, 1(Jun), pp.211-244.

\bibitem{b15}Fulcher, J., 2008. Computational intelligence: an introduction. In Computational intelligence: a compendium (pp. 3-78). Springer, Berlin, Heidelberg.

\bibitem{b16}Harvey, D.Y. and Todd, M.D. 2015. Automated feature design for numeric sequence classification by genetic programming. IEEE Transactions on Evolutionary Computation, 19(4), pp.474-489.

\bibitem{b17}Chatterjee, A. and Rong, M. 2018. Efficiency Analysis of Genetic Algorithm and Genetic Programming in Data Mining and Image Processing. Computer Vision: Concepts, Methodologies, Tools, and Applications. p.246.

\bibitem{b18}Tipping, M.E. and Faul, A.C. 2003. Fast marginal likelihood maximisation for sparse Bayesian models., AISTATS.

\bibitem{b19}Model, R.R., Square Adjusted R Square Std. Error of the Estimate, 1.

\bibitem{b20}Stijven, S., Vladislavleva, E., Kordon, A., Willem, L. and Kotanchek, M.E. 2016. Prime-Time: Symbolic Regression Takes Its Place in the Real World. In Genetic Programming Theory and Practice XIII (pp. 241-260). Springer, Cham.

\bibitem{b21}Bautu, E., Bautu, A. and Luchian, H. 2005. Symbolic regression on noisy data with genetic and gene expression programming. In Symbolic and Numeric Algorithms for Scientific Computing, 2005. SYNASC 2005. Seventh International Symposium.

\bibitem{b22}de Melo, V.V. and Banzhaf, W. 2015. Predicting high-performance concrete compressive strength using features constructed by kaizen programming. In Intelligent Systems pp. 80-85.

\bibitem{b23}de Melo, V.V. and Banzhaf, W. 2017. Improving the prediction of material properties of concrete using Kaizen Programming with Simulated Annealing. Neurocomputing, 246, pp.25-44.

\bibitem{b24}de Melo, V.V., 2016. Breast cancer detection with logistic regression improved by features constructed by Kaizen programming in a hybrid approach. In IEEE Evolutionary Computation Congress, pp. 16-23.

%

\bibitem{b25}dal Piccol Sotto, L.F. and de Melo, V.V., 2016. Studying bloat control and maintenance of effective code in linear genetic programming for symbolic regression. Neurocomputing, 180, pp.79-93.

\bibitem{b26}Sotto, L.F., Coelho, R.C. and de Melo, V.V., 2016, July. Classification of Cardiac Arrhythmia by Random Forests with Features Constructed by Kaizen Programming with Linear Genetic Programming. In Proceedings of the Genetic and Evolutionary Computation Conference 2016 (pp. 813-820). ACM.

\bibitem{b27}Arnaldo, I., Krawiec, K. and O'Reilly, U.M. 2014. Multiple regression genetic programming. In Proceedings of the 2014 Annual Conference on Genetic and Evolutionary Computation, pp. 879-886.

\bibitem{b28}Efron, B., Hastie, T., Johnstone, I. and Tibshirani, R. 2004. Least angle regression. The Annals of statistics, 32(2), pp.407-499.

\bibitem{b29}Moraglio, A., Krawiec, K. and Johnson, C.G. 2012. Geometric semantic genetic programming. In International Conference on Parallel Problem Solving from Nature, pp. 21-31, Springer, Berlin, Heidelberg.

\bibitem{b30}Castelli, M., Trujillo, L., Vanneschi, L. and Silva, S. 2015. Geometric semantic genetic programming with local search. In Proceedings of the 2015 Annual Conference on Genetic and Evolutionary Computation, pp. 999-1006.

\bibitem{b31}Christopher, M.B. 8th printing 2009. PATTERN RECOGNITION AND MACHINE LEARNING. Springer-Verlag New York.

\bibitem{b32}McDermott, J., et. al. ,2012. Genetic programming needs better benchmarks. In Proceedings of the 14th annual conference on Genetic and evolutionary computation pp. 791-798.

\bibitem{b33}Karaboga, D., Ozturk, C., Karaboga, N. and Gorkemli, B. 2012. Artificial bee colony programming for symbolic regression. Information Sciences, 209, pp.1-15.

\bibitem{b34}de Melo, V.V. and Banzhaf, W. 2016. Improving logistic regression classification of credit approval with features constructed by Kaizen programming. Genetic and Evolutionary Computation Conference Companion pp. 61-62.







\end{thebibliography}
\end{document}